\documentclass[letterpaper]{article} 
\usepackage{aaai24}  
\usepackage{times}  
\usepackage{helvet}  
\usepackage{courier}  
\usepackage[hyphens]{url}  
\usepackage{graphicx} 
\usepackage{natbib}  
\usepackage{caption} 
\usepackage{algorithm}
\usepackage{algorithmic}
\usepackage{sectsty}
\usepackage{caption} 
\usepackage{multirow}
\usepackage{booktabs}       

\usepackage{newfloat}
\usepackage{listings}
\DeclareCaptionStyle{ruled}{labelfont=normalfont,labelsep=colon,strut=off} 
\floatstyle{ruled}
\newfloat{listing}{tb}{lst}{}
\floatname{listing}{Listing}
\title{Rethinking Attention: Exploring Shallow Feed-Forward Neural Networks as an Alternative to Attention Layers in Transformers}
\author{
    Vukasin Bozic,
    Danilo Dordevic,
    Daniele Coppola,
    Joseph Thommes,
    Sidak Pal Singh
}
\affiliations{
    ETH Zurich \\
    \{vbozic, ddordevic, dcoppola, jthommes\}@student.ethz.ch,
    sidak.singh@inf.ethz.ch
}

\begin{document}

\maketitle

\begin{abstract}
\emph{
This work presents an analysis of the effectiveness of using standard shallow feed-forward networks to mimic the behavior of the attention mechanism in the original Transformer model, a state-of-the-art architecture for sequence-to-sequence tasks. We substitute key elements of the attention mechanism in the Transformer with simple feed-forward networks, trained using the original components via knowledge distillation. Our experiments, conducted on the IWSLT2017 dataset, reveal the capacity of these ”attentionless Transformers” to rival the performance of the original architecture. Through rigorous ablation studies, and experimenting with various replacement network types and sizes, we offer insights that support the viability of our approach. This not only sheds light on the adaptability of shallow feed-forward
networks in emulating attention mechanisms but also underscores their potential to streamline complex architectures for sequence-to-sequence tasks.}
\end{abstract}
\section{Introduction}
The seminal paper \cite{attention_is_all_you_need} which introduced the Transformer model has fundamentally altered the landscape of sequence-to-sequence modeling tasks. It set new benchmarks for language translation, measured by the BLEU score \cite{bleu}. The Transformer's attention mechanism enables the establishment of long-term dependencies in sequential data, allowing it to attend to every element in a sequence, a feat prior network architectures struggled to achieve without significant computational overheads.

Inspired by prior work \cite{ba_deep_2014}, \cite{urban_deep_2017} which explore the feasibility of training shallow feed-forward networks to emulate the behavior of deep convolutional networks with deep networks as teachers, we conduct a similar investigation on the original Transformer presented in \cite{attention_is_all_you_need}. Our focus is on language translation, utilizing the IWSLT2017 dataset \cite{cettolo-etal-2017-overview}. We aim to assess the extent to which standard shallow feed-forward networks can model attention mechanisms by substituting key attention components with feed-forward networks trained to replicate their behavior. \\

This work provides empirical evidence supporting the notion that shallow feed-forward networks can effectively learn the behaviors of Transformer attention modules and replace them without significantly impacting its overall performance. While it does not introduce a competitive advantage over established methods, it offers a conceptual analysis of existing techniques and potential alternatives.

\begin{figure}[H]
\includegraphics[width=1\columnwidth]{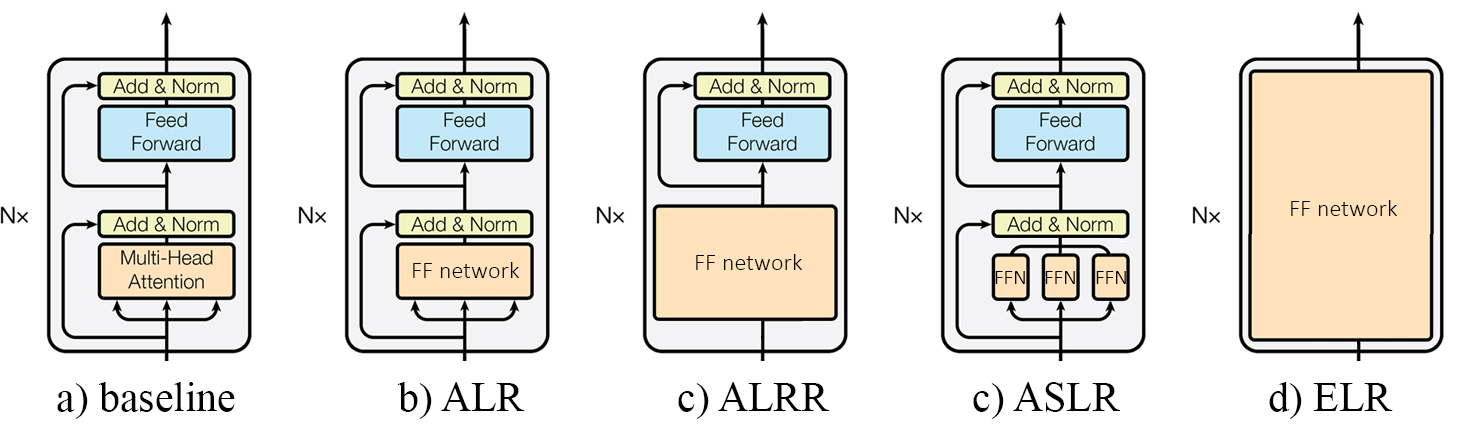} 
\caption{Different encoder self-attention replacement approaches presented.}
\label{fig: methods}
\end{figure}

\section{Models and Method}

The Transformer architecture is composed of stacked encoder and decoder blocks, which use attention to process input data. The encoder layer features one self-attention block, while the decoder layer encompasses both self-attention and cross-attention blocks, fusing the data processed by the encoder and itself. This model was used as the baseline, i.e. the teacher model, where the intermediate activations of its blocks were used for knowledge distillation \cite{hinton_distilling_2015} in the training of the feed-forward networks.

\textbf{Encoder self-attention replacement.} In the proposed approach, a thorough ablation study of the potential replacement methods was conducted. The experiments were done on self-attention layers in all 6 encoder blocks.

We introduced four different levels of abstraction for replacing the original encoder attention: \\
Attention Layer Replacement (\textbf{ALR}), Attention Layer with Residual Connection Replacement (\textbf{ALRR}), Attention Separate Heads Layer Replacement (\textbf{ASLR}) and  Encoder Layer Replacement (\textbf{ELR}), as depicted in Figure \ref{fig: methods}. Furthermore, all of these architectures were trained in 5 different sizes, ranging from "XS" to "L".

\textbf{Full Transformer attention replacement.} As ALR was found to be the most effective approach in the case of encoder attention replacement, featuring both high performance and a small number of parameters, the whole procedure was recreated for decoder self-attention and cross-attention replacement. This required adaptations of the previously introduced architectures, caused by different types of attention in the decoder. More details regarding the motivation and the choice of the replacement networks are given in Appendix A\ref{sec: App_A}, while implementation and training specifics of all of the FF replacement networks are provided in the Appendix B\ref{sec: App_B}.

\section{Results}
BLEU metric was used for evaluation purposes in this work, as it represents a standard metric for language translation tasks. The results for both encoder self-attention and full Transformer replacement studies span on 4 subsets of the IWSLT2017 dataset. Furthermore, BLEU scores relative to the baseline (vanilla Transformer) score of every experiment were calculated and then averaged over the datasets. Partial experimental results are presented in Figures \ref{fig: diff_methods} and \ref{fig: diff_places}, while the complete results are available in the Appendix C \ref{sec: App_C}.  We provide the implementation code on Github\footnote{https://github.com/vulus98/Rethinking-attention.git}.

\section{Discussion}
\begin{figure}[t]
\centering
\includegraphics[width=1\columnwidth]{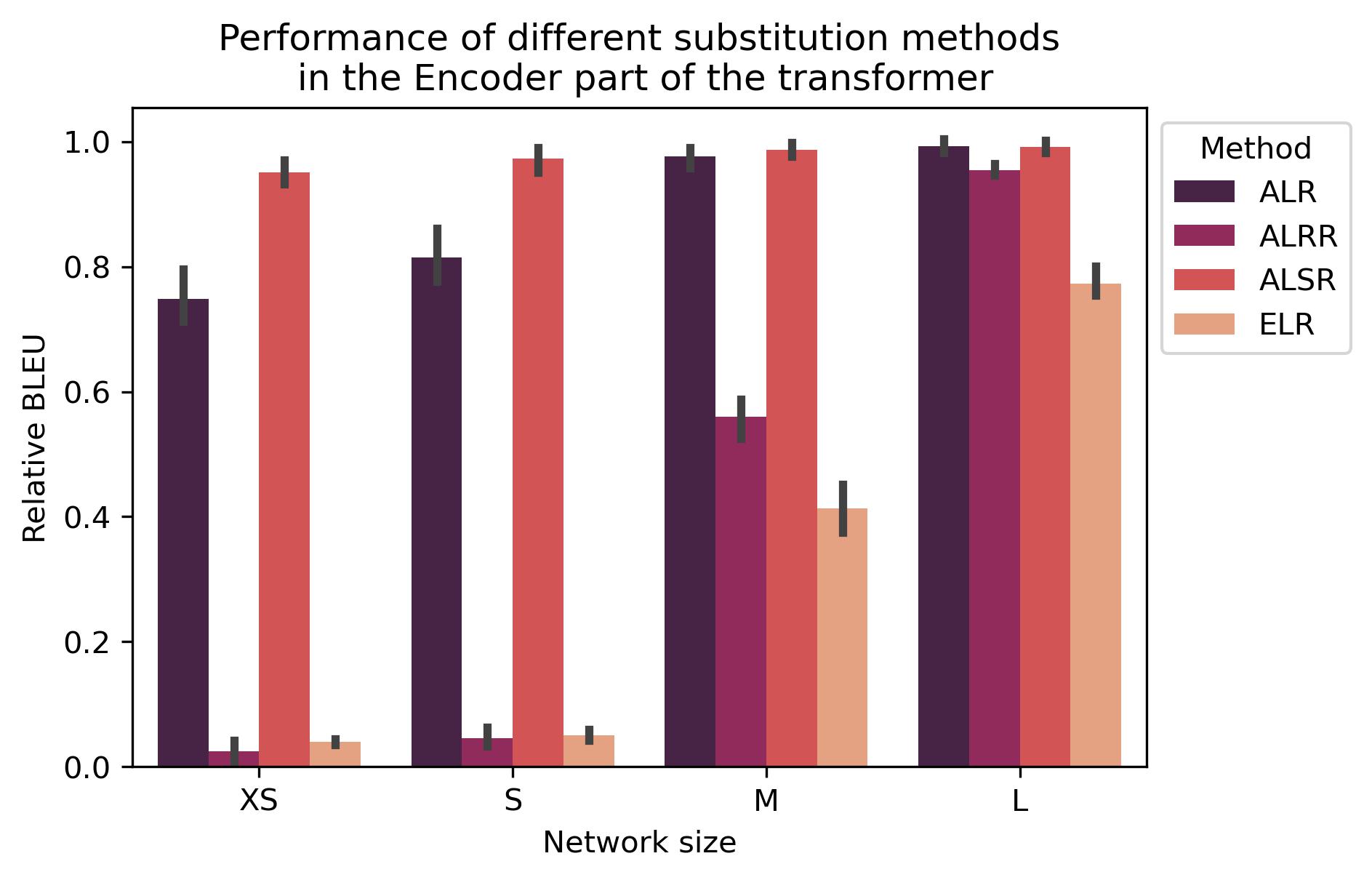} 
\caption{Relative BLEU scores [\%] (relative to the baseline Transformer), depending on the FF network size. Encoder self-attention is replaced using different replacement methods.}
\label{fig: diff_methods}
\end{figure}

In the case of encoder replacement, all of the proposed methods achieve competitive results compared to the baseline, as seen in Figure \ref{fig: diff_methods}. Out of the four approaches, ELR performs the worst, which is caused by the simplicity of the replacement model, which discards all of the encoder structures that aid training.

\begin{figure}[t]
\centering
\includegraphics[width=1\columnwidth]{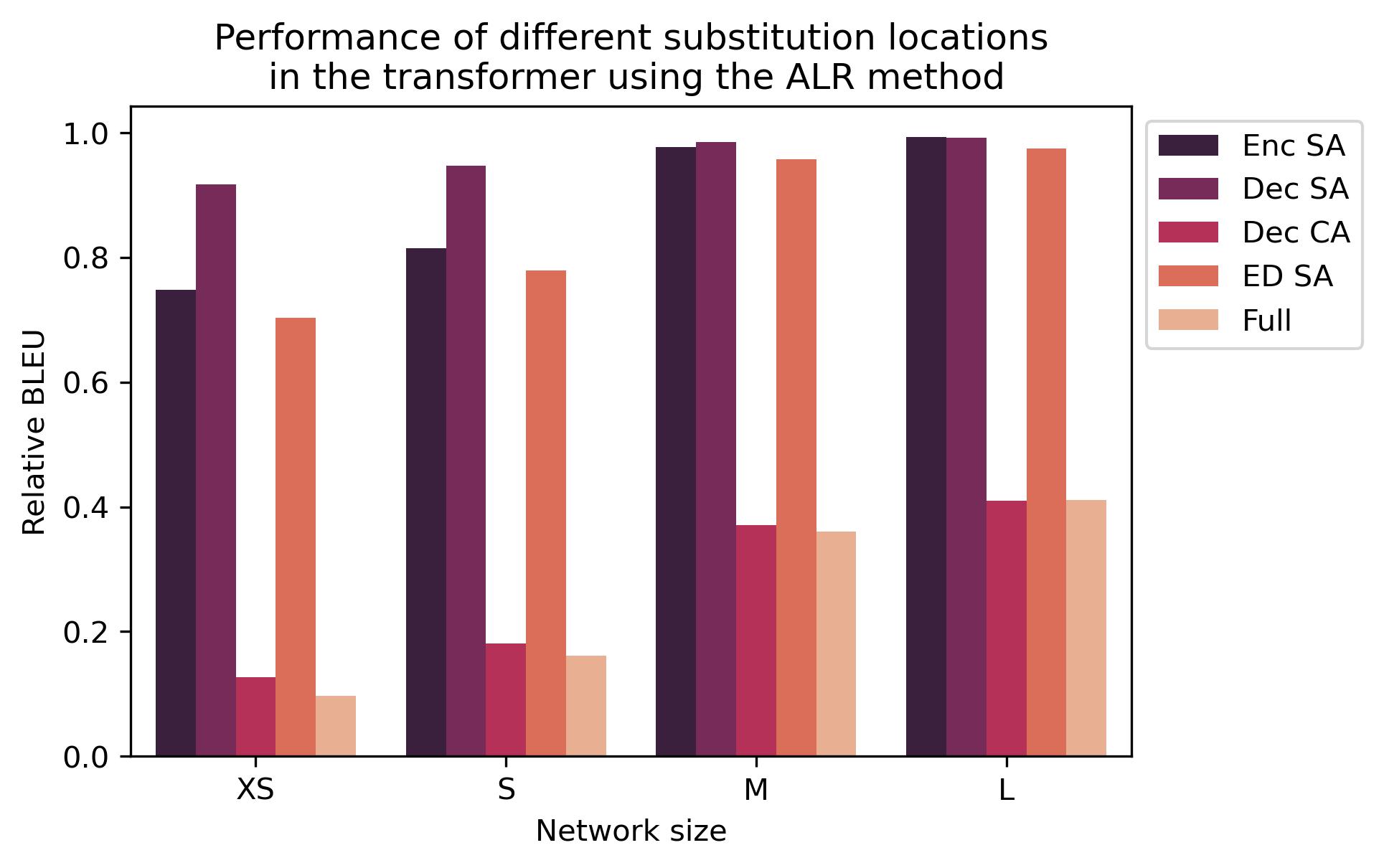} 
\caption{Relative BLEU scores [\%] (relative to the baseline), depending on the FF network size. ALR method is used to replace different attention parts of the transformer.}
\label{fig: diff_places}
\end{figure}

Furthermore, the full Transformer replacement approach, where only the ALR method is utilized, yielded results showcasing the potential of the feed-forward networks to successfully replicate the decoder self-attention behavior, while the performance on decoder cross-attention is comparatively worse, as presented in Figure \ref{fig: diff_places}. The potential reason for this behaviour could be the lack of the expressiveness of the feed-forward network needed to describe the more complex mapping and interaction between sequences used in the cross-attention block, which also influences final evaluation scores for the fully "attentionless" Transformer.

However, all of the replacement approaches come at a significant cost of having more parameters. Another downside of our replacement of the attention with a fixed-size feed-forward network is the imminent lack of flexibility of the model in terms of the length of sequences the model can operate with.

\section{Conclusion}
Empirical evidence suggests that the proposed approaches are capable of achieving comparable performance to that of the original Transformer, demonstrating that Transformers do not necessarily need to have attention. These conclusions also point out the deficiencies of the current optimization methods, which are not able to train these "attentionless Transformers" from scratch but need more advanced techniques, such as knowledge distillation to converge into desired parameter configurations. This conclusion emphasizes that with the advancements in optimization techniques, less specialized architectures such as feed-forward networks could be used for advanced tasks, currently reserved for highly specialized architectures.

\section{Acknowledgements}
We would like to express our sincere gratitude to the Data Analytics lab of ETH Zurich for providing the necessary resources and support during the course of this project; the collaborative and enriching environment of the lab significantly contributed to the success of this research, and we are truly thankful for their invaluable assistance. Additionally, we extend our heartfelt thanks to G-research for their generous funding, which made it possible for us to attend the conference and present this paper.

\onecolumn

\appendix
\begin{centering}
    \section{\Huge Appendix}
\end{centering}

\section{Appendix A: Choice of the replacement networks} \label{sec: App_A}
Here we provide a more detailed description of the integration of the feed-forward (FF) networks into the Transformer architecture, and subsequent replacement of the attention modules. All of the replacement methods are based on replacing different sections of the attention module:
\begin{itemize}
    \item \textbf{Attention Layer Replacement (ALR)}: replaces only the multi-head attention (MHA) block with an FF network, keeping the residual connection and layer normalization intact.
    \item \textbf{Attention Layer with Residual Connection Replacement (ALRR)}: The MHA module, along with the residual connection is replaced by the FF network. This approach effectively removes the residual connection when the FF Network is substituted in the Transformer. 
    \item \textbf{Attention Separate heads Layer Replacement (ASLR)}: As a variant of the ALR, this method replaces every single head of the MHA module with a separate FF network.
    \item \textbf{Encoder Layer Replacement(ELR)}: Replaces the Encoder layer entirely.
\end{itemize}
Following is a description, motivation, and features of each of the aforementioned methods.
ALR and ALRR approaches were designed in such a way as to decouple the effects and benefits of the attention layer from the benefits brought by the residual connection. Furthermore, mimicking the behavior of each separate head, featured in the ASLR approach, makes the FF replacement more similar to the original MHA method. Finally, as the highest level of abstraction, the whole encoder block is replaced with an FF Network in the ELR method. This essentially upends the original encoder architecture, turning it into a sequence of FF networks - one for each block of the encoder. By inspecting all of the levels of substitution through experiments, we obtained a well-grounded ablation study on the possible attention mechanism replacements.
\newline

As a replacement module in all of these cases, the simple, shallow one-hidden-layer FF network was used. For all of the aforementioned approaches, FF networks of different sizes were devised, as given in Table \ref{table:num-params-method}. The original number of parameters for the attention layer (60,000 parameters in our case) is mostly exceeded by the replacement networks, mainly due to fixed-size inputs and outputs, and the processing format demanded by the FF network.

\begin{table}[ht]
\centering
    \begin{tabular}{lllll}
    \toprule
    & XS & S & M & L \\
    \midrule
    ALR & \multirow{3}{*}{320K} & \multirow{3}{*}{640K} & \multirow{3}{*}{10M} & \multirow{3}{*}{41M} \\
    ALRR & & & &\\
    ELR & & & &  \\
    \midrule
    ASLR & 290K & 1.5M & 11.5M & 46M \\
    \bottomrule
  \end{tabular}
  \caption{Number of parameters of the proposed architectures. The number featured is a number of parameters for a single FF network in one layer. Further scaling of these networks brought no improvement in the BLEU score. The number of parameters for the ASLR method are presented on a per-attention-head basis.}
\label{table:num-params-method}
\end{table}

\section{Appendix B: Implementation details} \label{sec: App_B}

The initial point of our procedure is the training of the vanilla Transformer model, which consists of six encoders and six decoders. To reduce training times and make testing faster, we reduced the embedding size from 512 to 128. These changes did not drop the overall score too much below the original BLEU score but resulted in significantly lower computational power demands. This Transformer model was then used as a teacher model for training the feedforward networks.

\begin{figure*}[t]
\centering
\includegraphics{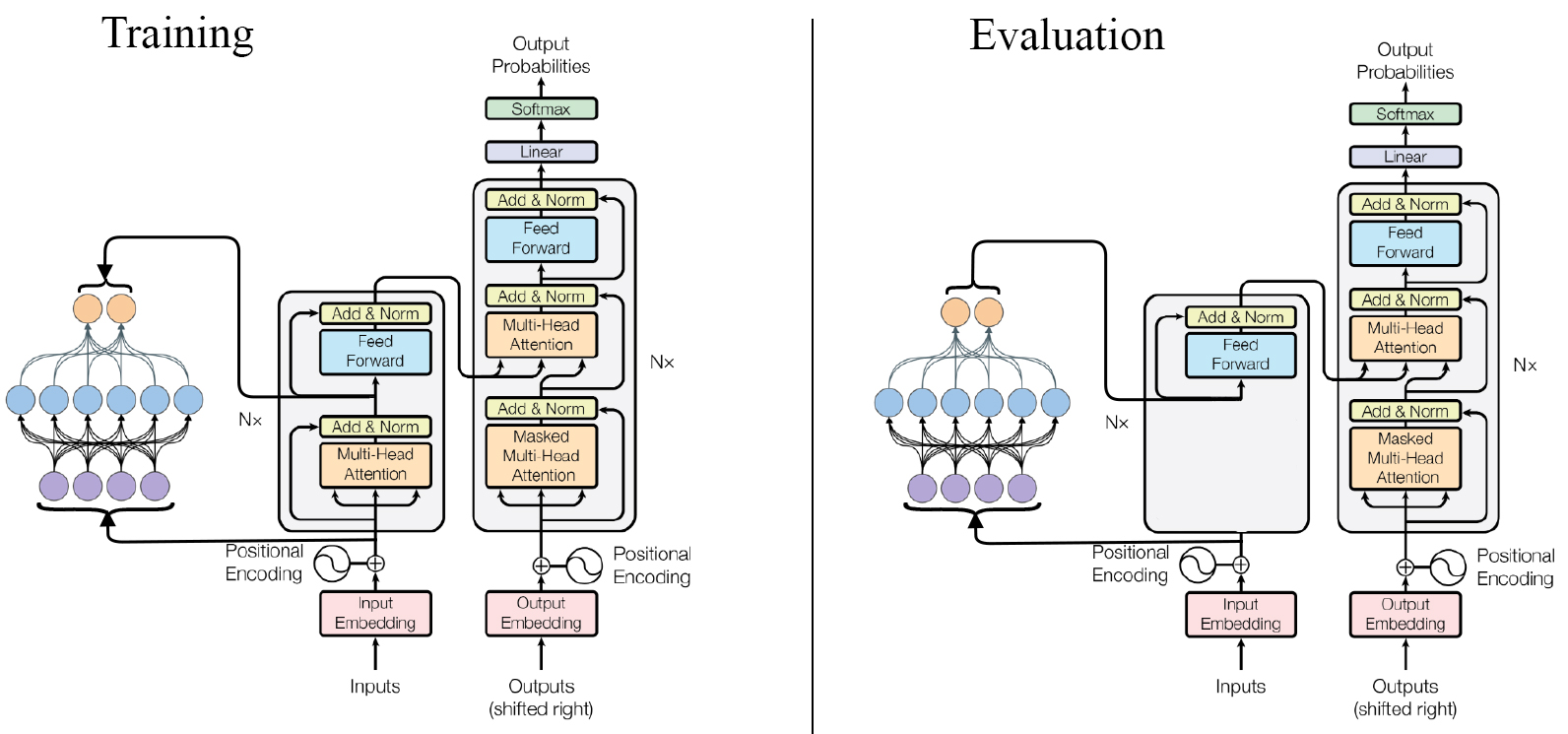}
\caption{Illustration of the training and evaluation cycles of the ALRR method in the encoder self-attention. Replacement in the self-attention and cross-attention layer is analogous. Other replacement methods follow the same principle, with the difference that the input data and teacher labels are taken from the different blocks of the encoder, depending on their structure.}
\label{fig: train_vs_eval}
\end{figure*}

In Figure \ref{fig: train_vs_eval}, the essence of our training and evaluation method is presented, on the example of ALRR substitution, while other approaches are analogous. As the initial step, intermediate activations (input-output pairs) are extracted from the trained Transformer and used as training data for the desired Feed-Forward replacement network. After that, they have to be additionally adapted, as described below.  \\

\begin{figure*}[t]
\centering
\includegraphics[scale=0.5]{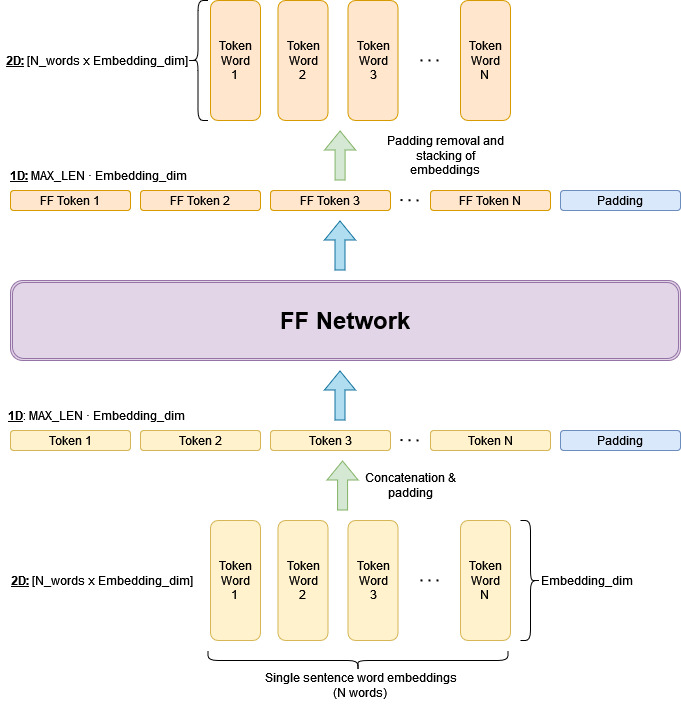}
\caption{Illustration of the necessary data preprocessing and post-processing before and after propagation through the Feed-forward network. }
\label{fig: ff_method}
\end{figure*}

\textbf{Necessary data transformations} Every attention layer transforms the input word representations of a sentence into a linear combination of values extracted by the input representation. To mimic this behavior, the FF network takes in the concatenated word representations of a sentence as input and produces updated word representations as output in a single pass. In order to handle the input sentences of varying lengths, we have decided to pad all sentences to a maximum fixed length and mask the padded values with zeros to prevent them from influencing the model's inference.  This process is illustrated in Figure \ref{fig: ff_method}.

We also adopted a fixed upper bound to sentence length, which we set to 50. This limit consequently limited the size of our replacement FF networks. Since 96\% of the samples in the datasets are of length 50 or less, the datasets were not significantly shrunk.

After the successful training of the networks, they are inserted in a Transformer architecture, replacing the now redundant layers, and the evaluation is run, as shown in Figure \ref{fig: train_vs_eval}.

\textbf{Decoder replacement networks} The main distinction between the replacement method used for the self-attention in the encoder and the one used for the self-attention in the decoder is that the FF replacement networks in the decoder output the word embeddings one at a time, following the original masked self-attention. The network processes each word individually by feeding the entire sentence representation through the network and masking the representation of words that come after the word is processed. This is done to take into account the concept of causality, where only the previous words in the sentence can affect the meaning of the current word.

Cross-attention in the decoder accepts both word representations from the encoder and decoder layers, integrating them together. To replicate this process, word representations from both encoder and decoder were concatenated together and padded, having the input size doubled in comparison to the self-attention replacement networks. Output dimensions stayed the same, again following the initial cross-attention design.

\textbf{FF networks training details} Every approach and every FF network size demanded training 6 independent networks for each of the self-attention or the cross-attention blocks.
The networks were trained for 20 epochs using the Adam as the optimizer of choice. The learning rate was set to 0.001, while the batch size was 1400. Training settings were kept the same for all networks. \\

\section{Appendix C: Datasets and Results} \label{sec: App_C}
For all of the testing procedures, IWSLT2017 dataset was used. It provides multiple subsets for language translations, among which we used French-English (F2E), English-French (E2F), German-English (G2E), and English-German (E2G) subsets. On average, these subsets consist of 200000 training and 1000 test sentences, obtained after shrinking the maximum size of the sentence to 50 words, as explained above.

The metric used for all of the tests was the BLEU score, as it provides an intuitive comparison to human-provided translation, puts the translation quality on a normalized 0-1 scale, and represents the main testing metric for many similar works in this field.

In the abstract we provided results averaged over all 4 datasets, proving the versatility and robustness of the proposed methods. Raw results are shown in Table \ref{table:results}, while the scores presented in the abstract were relative with respect to the baseline score of the original Transformer, on each respective dataset. The baseline scores are presented in Table \ref{table:baseline-scores}. 

We have replaced the attention mechanism in different ways in the encoder, decoder, and both. Results obtained during extensive experimentation suggest that the only obstacle to completely replacing the attention mechanisms in the Transformer is cross-attention. As evident from the table \ref{table:results}, BLEU scores are significantly lower if the cross-attention mechanism is replaced by FF networks. We tested the replacement in three ways: only in the decoder cross-attention, only in the encoder and decoder (without replacing the cross-attention), and full replacement. In all cases where the decoder cross-attention was substituted for the FF networks, the final model performed significantly worse, regardless of whether other parts were substituted.

This suggests that the proposed shallow networks were not able to capture the more intricate and complex interactions between the differing sequences that enter the cross-attention mechanism. On the other hand, the self-attention was modeled and learned successfully.  

\section{Appendix D: Future work} \label{sec: App_D}
By matching the performance of the original Transformer, it is highly probable that the further optimization of the FF networks' hyperparameters using advanced parameter search (e.g. using Bayesian optimization \cite{bayes_optimization}) could yield even better results in terms of translation quality and possibly even enable the usage of smaller FF networks for the replacement, as the size of the networks represents one of the major bottlenecks for the deployment of these 'attentionless' Transformers in practice.

Furthermore, another potential direction lies in the training of more complex FF networks for the purpose of modeling the cross-attention module of the decoder, as the current shallow network shows that, in contrast to self-attention which they can learn successfully, cross-attention proves to be more challenging due to its complexity.

\begin{table}[ht]
\centering
    \begin{tabular}{l|llll}
    \toprule
     & E2G & G2E & E2F & F2E \\
     \midrule
     Transformer & 0.257 & 0.324 & 0.276 & 0.292 \\
    \bottomrule
  \end{tabular}
  \caption{Baseline transformer BLEU scores on all 4 four language-translation datasets.}
\label{table:baseline-scores}
\end{table}

\begin{table}[H]
\begin{centering}
\renewcommand{\arraystretch}{1.1} 
\begin{tabular}{|l|l|l|llll|}
    \hline
     & & & E2G & G2E & E2F & F2E \\
    \hline
    \multirow{20}{*}{ALR} & \multirow{4}{*}{\shortstack{Enc\\SA}} & XS & 0.180 & 0.235 & 0.226 & 0.218 \\
     & & S & 0.196 & 0.257 & 0.244 & 0.240 \\
     & & M & 0.245 & 0.320 & 0.275 & 0.284 \\
     & & L & 0.252 & 0.327 & 0.276 & 0.288 \\
    \cline{2-7}
     & \multirow{4}{*}{\shortstack{Dec\\SA}} & XS & 0.227 & 0.305 & 0.251 & 0.273 \\
     & & S & 0.240 & 0.313 & 0.255 & 0.281 \\
     & & M & 0.252 & 0.322 & 0.267 & 0.290 \\
     & & L & 0.253 & 0.323 & 0.273 & 0.291 \\
    \cline{2-7}
     & \multirow{4}{*}{\shortstack{Dec\\CA}} & XS & 0.035 & 0.036 & 0.035 & 0.039 \\
     & & S & 0.054 & 0.058 & 0.042 & 0.053 \\
     & & M & 0.104 & 0.125 & 0.089 & 0.108 \\
     & & L & 0.115 & 0.130 & 0.109 & 0.115 \\
    \cline{2-7}
     & \multirow{4}{*}{\shortstack{E-D\\SA}} & XS & 0.163 & 0.222 & 0.219 & 0.204 \\
     & & S & 0.187 & 0.247 & 0.235 & 0.227 \\
     & & M & 0.244 & 0.313 & 0.265 & 0.277 \\
     & & L & 0.246 & 0.321 & 0.270 & 0.284 \\
    \cline{2-7}
     & \multirow{4}{*}{Full} & XS & 0.026 & 0.027 & 0.026 & 0.032 \\
     & & S & 0.041 & 0.053 & 0.048 & 0.044 \\
     & & M & 0.102 & 0.122 & 0.083 & 0.107 \\
     & & L & 0.105 & 0.134 & 0.117 & 0.116 \\
    \hline
    \multirow{4}{*}{ALRR} & \multirow{4}{*}{\shortstack{Enc\\SA}} & XS & 0.013 & 0.010 & 0.003 & 0.001 \\
     & & S & 0.018 & 0.013 & 0.008 & 0.012 \\
     & & M & 0.158 & 0.181 & 0.153 & 0.150 \\
     & & L & 0.243 & 0.315 & 0.263 & 0.276 \\
     \hline
    \multirow{4}{*}{ASLR} & \multirow{4}{*}{\shortstack{Enc\\SA}} & XS & 0.245 & 0.319 & 0.257 & 0.272 \\
     & & S & 0.250 & 0.323 & 0.260 & 0.285 \\
     & & M & 0.251 & 0.326 & 0.269 & 0.289 \\
     & & L & 0.252 & 0.326 & 0.271 & 0.290 \\
     \hline
    \multirow{4}{*}{ELR} & \multirow{4}{*}{\shortstack{Enc\\SA}} & XS & 0.012 & 0.010 & 0.011 & 0.011 \\
     & & S & 0.016 & 0.015 & 0.132 & 0.012 \\
     & & M & 0.116 & 0.120 & 0.124 & 0.110 \\
     & & L & 0.194 & 0.248 & 0.225 & 0.219 \\
     \hline
\end{tabular}
\caption{BLEU scores of all proposed methods, in all sizes and over all used datasets. Abbreviations were used for clarity purposes, and are explained in the following text: "Enc" stands for encoder, "Dec" stands for decoder, "SA" stands for self-attention, "CA" stands for cross-attention, and E-D stands for encoder and decoder.}
\label{table:results}
\end{centering}
\end{table}


\begin{thebibliography}{7}
\providecommand{\natexlab}[1]{#1}

\bibitem[{Ba and Caruana(2014)}]{ba_deep_2014}
Ba, L.~J.; and Caruana, R. 2014.
\newblock Do {Deep} {Nets} {Really} {Need} to be {Deep}?
\newblock ArXiv:1312.6184 [cs].

\bibitem[{Cettolo et~al.(2017)Cettolo, Federico, Bentivogli, Niehues, St{\"u}ker, Sudoh, Yoshino, and Federmann}]{cettolo-etal-2017-overview}
Cettolo, M.; Federico, M.; Bentivogli, L.; Niehues, J.; St{\"u}ker, S.; Sudoh, K.; Yoshino, K.; and Federmann, C. 2017.
\newblock Overview of the {IWSLT} 2017 Evaluation Campaign.

\bibitem[{Hinton, Vinyals, and Dean(2015)}]{hinton_distilling_2015}
Hinton, G.; Vinyals, O.; and Dean, J. 2015.
\newblock Distilling the {Knowledge} in a {Neural} {Network}.
\newblock ArXiv:1503.02531 [cs, stat].

\bibitem[{Papineni et~al.(2002)Papineni, Roukos, Ward, and Zhu}]{bleu}
Papineni, K.; Roukos, S.; Ward, T.; and Zhu, W.-J. 2002.
\newblock BLEU: A Method for Automatic Evaluation of Machine Translation.

\bibitem[{Snoek, Larochelle, and Adams(2012)}]{bayes_optimization}
Snoek, J.; Larochelle, H.; and Adams, R.~P. 2012.
\newblock Practical Bayesian Optimization of Machine Learning Algorithms.
\newblock arXiv:1206.2944.

\bibitem[{Urban et~al.(2017)Urban, Geras, Kahou, Aslan, Wang, Caruana, Mohamed, Philipose, and Richardson}]{urban_deep_2017}
Urban, G.; Geras, K.~J.; Kahou, S.~E.; Aslan, O.; Wang, S.; Caruana, R.; Mohamed, A.; Philipose, M.; and Richardson, M. 2017.
\newblock Do {Deep} {Convolutional} {Nets} {Really} {Need} to be {Deep} and {Convolutional}?
\newblock ArXiv:1603.05691 [cs, stat].

\bibitem[{Vaswani et~al.(2017)Vaswani, Shazeer, Parmar, Uszkoreit, Jones, Gomez, Kaiser, and Polosukhin}]{attention_is_all_you_need}
Vaswani, A.; Shazeer, N.; Parmar, N.; Uszkoreit, J.; Jones, L.; Gomez, A.~N.; Kaiser, L.; and Polosukhin, I. 2017.
\newblock Attention Is All You Need.
\newblock arXiv:1706.03762.

\end{thebibliography}
\end{document}